# Efficient Tracking of a Moving Object using Inter-Frame Coding

[1] Shraddha Mehta
[1] Student, RK. University, Rajkot, Gujrat, India

[2] Vaishali Kalariya
[2] Assistant Professor, RK. University, Rajkot, Gujrat, India

*Abstract*—Video surveillance has long been in use to monitor security sensitive areas such as banks, department stores, highways, crowded public places and borders. The advance in computing power, availability of large-capacity storage devices and high speed network infrastructure paved the way for cheaper, multi sensor video surveillance systems. Traditionally, the video outputs are processed online by human operators and are usually saved to tapes for later use only after a forensic event. The increase in the number of cameras in ordinary surveillance systems overloaded both the human operators and the storage devices with high volumes of data and made it infeasible to ensure proper monitoring of sensitive areas for long times. In order to filter out redundant information generated by an array of cameras, and increase the response time to forensic events, assisting the human Operators with identification of important events in video by the use of "smart" video surveillance systems has become a critical requirement. The making of video surveillance systems "smart" requires fast, reliable and robust algorithms for moving object detection, classification, tracking and activity analysis.

## I. INTRODUCTION

The system is initialized by feeding video imagery from a static camera monitoring a site. Most of the methods are able to work on both colour and monochrome video imagery. The first step of our approach is distinguishing foreground objects from stationary background. To achieve this, we use a combination of adaptive background subtraction and low-level image post-processing methods to create a foreground pixel map at every frame. We then group the connected regions in the foreground map to extract individual object features such as bounding box, Area, centre of mass and colour histogram.

The object tracking algorithm utilizes extracted object features together with a correspondence matching scheme to track objects from frame to frame. The colour histogram of an object produced in previous step is used to match the correspondences of objects after an occlusion event. The output of the tracking Step is object trajectory information which is used to calculate direction and speed of the objects in the scene.

The detection method is detecting the foreground pixels by using the background model and the current image from video. This pixel-level detection process is dependent on the background model in use and it is used to update the background model to adapt to dynamic scene changes. Also, due to camera noise or environmental effects the detected foreground pixel map contains noise. Pixel-level post-processing operations are performed to remove noise in the foreground pixels.

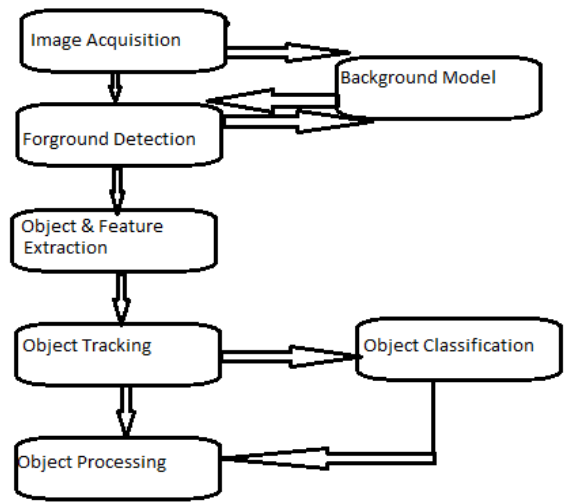

**Figure 1.1** Object Tracking Block - Diagram

## II. ALGORITHM

1. Initialization:

2. Calculate the target model $q$ and initialize the position $y_0$ of the target candidate model in the previous frame.

3. Initialize the iteration number $k \leftarrow 0$.

4. Calculate the target candidate model $p\ y_0$ in the current frame.

5. Calculate the new position $y_1$ of the target candidate model.

6. Let $d \leftarrow \|y_1-y_0\|$, $y_0 \leftarrow y_1$. Set the error threshold $\varepsilon$ (default 0.1) and the maximum Iteration number.

7. Estimate the width, height and orientation from the target candidate model.

8. Estimate the initial target candidate model for next frame.

## III. BACKGROUND SUBTRACTION

Let $I\eta(x)$ represent the grey-level intensity value at pixel position (x) and at time instance n of video image sequence I which is in the range [0, 255]. Let $B\eta(X)$ be the corresponding background intensity value for pixel position (x) estimated over time from video images I0 through $I\eta-1(x)$. As the generic background subtraction scheme suggests, a pixel at position (x) in the current video image belongs to foreground if it satisfies:

$$|II\eta(x) - B\eta(x)| > T\eta(x)$$

Where $T\eta(x)$ an adaptive threshold value is estimated using the image sequence I0 through $I\eta-1(x)$. The Equation is used to generate the foreground pixel map which represents the foreground regions as a binary array where a 1 corresponds to a foreground pixel and a 0 stands for a background pixel.

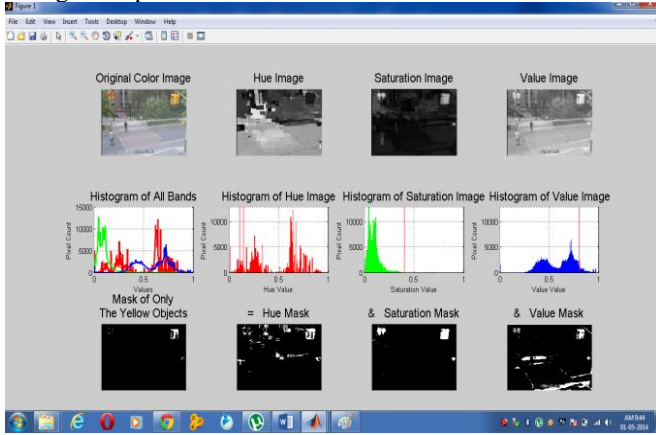

**Figure 3.1** Background Subtraction

## IV. CORRESPONDENCE & HISTOGRAM-BASED OBJECT MATCHING

The first step in our object tracking algorithm is matching the objects ($O\rho$'s) in previous image $I\eta-1(x)$ to the new objects ($Oi$'s) detected in current image (In). We store the matching of objects in a bi-partite graph $G(m,n)$. In this graph, vertices represent the objects (one vertex partition represents previous objects, $C\rho$ $O\rho$'s and the other partition represents new objects, $Oi$'s) and edges represent a match between two objects. In $G(m,n)$, m is the size of the partition for previous objects, and n is the size of the partition for the new objects other words, two objects with center of mass points $FC = \{c1, c2, c3....cn\}$ $C\rho$ and $Ci$ are close to each other if the following is satisfied:

$$Dist(C\rho, Ci) < \lambda$$
$$Dist(C\rho, Ci) = \sqrt{(Xc\rho - Xci)2} - \sqrt{(Yc\rho - Yci)2}$$

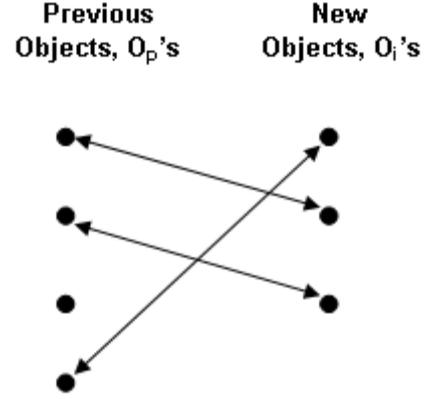

**Figure 4.1**: Sample object matching graph.

The distance dab between two normalized color histograms Ha and Hb with N bins are calculated by using the L1 metric as follows:

$$Dab = \sum_i^n \lfloor Ha(i) - Hb(i) \rfloor$$

$$Dtotal = Dupperhistogram + Dlowerhistogram$$

$$DS[i] = DS[i * \frac{N}{C}], \forall i \in [1........C]$$

$$DS[i] = \frac{DS[i]}{\sum_1^n DS[i]}$$

Where, $DS[i]$ is normalized to have integral unit area.

## V. EXPERIMENTAL RESULTS

**Colour Detection**: We decided to represent the fire colour cloud by using a mixture of Gaussians in RGB colour space. In this approach, the sample set of fire colours $FC = \{c1, c2, c3....cn\}$ is considered as a pixel process and a Gaussian mixture model with N(= 10) Gaussian distributions is initialized by using these samples. In other words, we represent the point cloud of fire coloured pixels in RGB space by using N spheres whose union almost covers the point cloud.

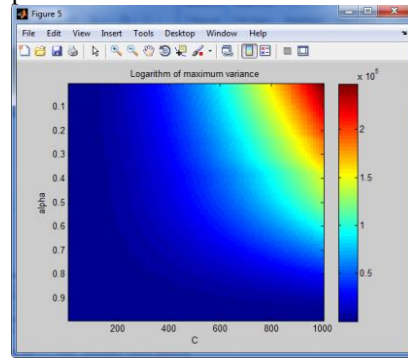

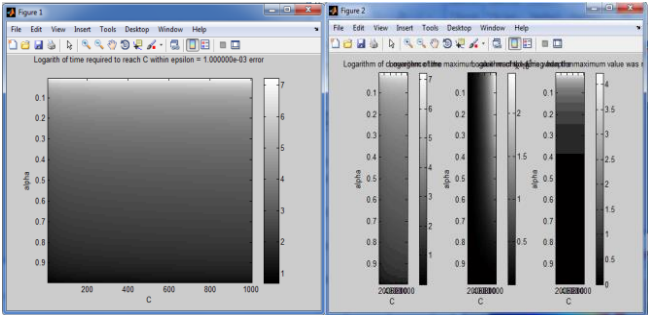

**Figure 5.1**: Threshold Value of Colours

| | |
|---|---|
| ALPHA | 0.0200 |
| RHO | 0.0100 |
| DEVIATION_SQ_THRESHOLD | 49 |
| INIT_VARIANCE | 3 |
| INIT_MIXPROP | 1.0000e-0.5 |
| BACKGROUND_THRESHOLD | 0.9000 |
| COMPONENT_THRESHOLD | 10 |

**Table 5.1**: Initial Threshold Value

$$p(d) = \{p(Y = y) * (X = x)\}$$

$$p(c) = \{(p(R = r)\Im p(G = g)\Im p(B = b)\}$$

Where r=Intensity Value of red

g= Intensity Value of green

b= Intensity Value of blue

x= Intensity Value of any X colour

y= Intensity Value of any Y colour

**Demo-**

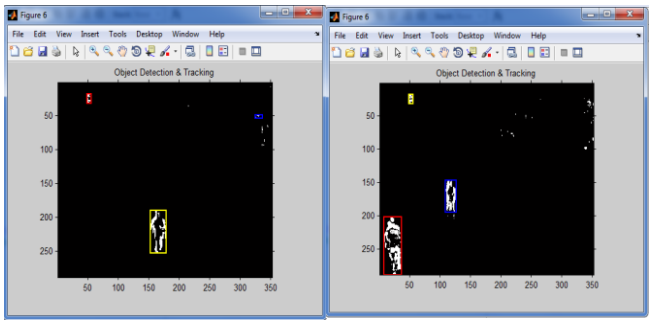

Figure (1)       Figure (2)

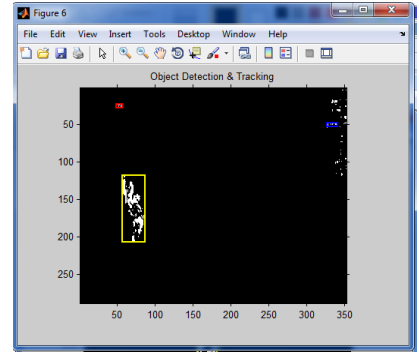

Figure (3)

**Figure 5.3**: Human Tracking Demo (1) Only one person in frame, it shows with yellow colour bounding box 2) While other person comes in frame at that time first person changes it colour from yellow to green bounding box and second person treated with red colour bounding box.(3) While two persons (objects) collide with each other it treated as a noise with yellow colour bounding box.

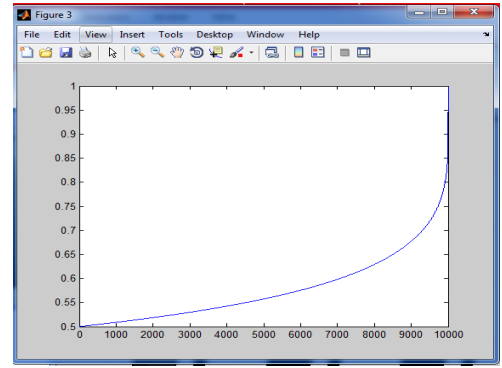

**Figure 5.4:** Pictorial Graphical Representation of Human Tracking Demo Program

V. CONCLUSION

This paper has focused on an important task to detect and track objects ahead with video camera. Our approach is mainly based on motion information. Several general features that characterize the objects ahead are robustly extracted in the video. Automatic tracking is a very interesting research area that can lead to numerous intelligent applications. Our work focused on setting up a system that may be used for intelligent applications in video surveillance scenarios. Real Time videos are more challenging because of illumination changes, non-static backgrounds, and occlusion. The high amount of noise and uncertainty observed in outdoor sequences makes tracking in these sequences a difficult problem to solve. We have designed a tracking system that is able to detect and track moving objects in real video. After setting up a basic system, we were able to bring significant improvements in the tracking by use of new algorithms. Thus, we were successful in both our goals of establishing a base system that can serve as a platform for future Automatic Video Surveillance research of developing new algorithms to further the state of research in the field.

## REFERENCES

[1] Luigi Di Stefano, Enrico Viarani,"Vehicle Detection and Tracking Using the Block Matching Algorithm",*in IEEE*-2013

[2] R. Bodor, B. Jackson, and N. Papanikolopoulos. Vision-based human tracking and activity recognition. *In IEEE*, -2012.

[3] M. Betke and H. Nguyen, "Highway scene analysis from a moving vehicle under deduced visibility conditions," in *Proc. IEEE Intellectual. Vehicle*, 2012, pp. 131–136.

[4] A. Cavallaro and F. Ziliani.," Image Analysis for Advanced Video Surveillance",*in IEEE*-2011

[5] D. Alonso, L. Salgado, and M. Nieto, "Robust vehicle detection through multidimensional classification for on board video based systems," in *Procedural IEEE ICIP*, Sep. 2011, vol. 4, pp. 321–324.

[6] L. Gao, C. Li, T. Fang, and Z. Xiong, "Vehicle detection based on color and edge information," in *Proc. Image Anal. Recog.*, vol. 5112, *Lect. Notes Comput. Sci.*, 2010, pp. 142–150.

[7] H.T. Chen, H.H. Lin, and T.L. Liu. "Multi-object tracking using dynamical graph matching." *In Proc. of IEEE Computer Society Conference on Computer Vision and Pattern Recognition*, pages 210–217, 2009..

[8] W. Zhang, X. Z. Fang, and X. K. Yang, "Moving vehicles segmentation based on Bayesian framework for Gaussian motion model," *Pattern Recognit. Lett.*, vol. 27, no. 9, pp. 956–967, Jul. 2009.

[9] C. Demonceaus, A. Potelle, and D. Kachi-Akkouche, "Obstacle detection in a road scene based on motion analysis," *IEEE Trans. Veh. Technol.*, vol. 53, no. 6, pp. 1649–1656, Nov. 2008.

[10] S. Sivaraman and M. Trivedi, "A general active-learning framework for on-road object recognition and tracking," *IEEE Trans. Intell. Transp. Syst.*, vol. 11, no. 2, pp. 267–276, Jun. 2008.

[11] N. Ghosh and B. Bhanu, "Incremental unsupervised three-dimensional vehicle model learning from video," *IEEE Trans. Intell. Transp. Syst.*, vol. 11, no. 2, pp. 423–440, Jun. 2006.

[12] R. Cutler and L.S. Davis. Robust real-time periodic motion detection, analysis and applications. *In IEEE Transactions on Pattern Analysis and Machine Intelligence*, volume 8, pages 781–796, 2004.

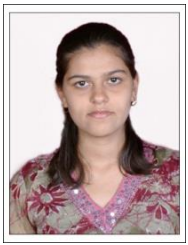

Shraddha Mehta received her B.E. degree in Computer Science Engineering from Gujarat Technological University, India, in 2012, and pursuing M.Tech. degree in Computer Engineering from RK University, Rajkot, India, in 2014, She was a lecturer with Department of Computer Engineering, RK University, in 2012 to 2013.Her research interests includes with Image Processing. At present, She is engaged in Real Time Object Tracking Based On Inter-Frame Coding.

Vaishali Kalariya received her M.E degree in Computer Engineering and pursuing PHD from RK University, Rajkot, Gujarat, India. She is assistant Professor with Department of Computer Engineering, RK University. Her research interests includes with Image Processing.